\pdfoutput=1

\documentclass[11pt]{article}

\usepackage[final]{coling}
\usepackage[normalem]{ulem} 

\usepackage{times}
\usepackage{latexsym}

\usepackage[T1]{fontenc}

\usepackage[utf8]{inputenc}

\usepackage{amssymb}
\usepackage{bbm}
\usepackage{bm}
\usepackage{titlesec}
\usepackage{multirow}
\usepackage{mathtools}
\usepackage{colortbl}
\usepackage{float}
\usepackage{xcolor}
\usepackage{adjustbox}
\usepackage{enumitem}
\usepackage{arydshln}
\usepackage[capitalize]{cleveref}

\usepackage{booktabs}
\usepackage[font=small,labelfont=bf,tableposition=top]{caption}
\DeclareCaptionLabelFormat{andtable}{#1~#2  \&  \tablename~\thetable}
\usepackage{microtype}

\usepackage{tikz}
\newcommand*\circled[1]{\tikz[baseline=(char.base)]{
            \node[shape=circle,draw,inner sep=1pt] (char) {#1};}}
\NewDocumentCommand{\ken}
{ mO{} }{\textcolor{purple}{\textsuperscript{\textit{Ken}}\textsf{\textbf{\small[#1]}}}}

\usepackage{microtype}

\usepackage{inconsolata}

\usepackage{graphicx}

\usepackage{enumitem}
\usepackage{algorithm}
\usepackage{tabularx}
\usepackage{subcaption}
\usepackage[most]{tcolorbox}
\definecolor{tablerow1}{RGB}{225,217,205}
\definecolor{tablerow2}{RGB}{236,229,221}

\definecolor{RoseQuartzBg}{HTML}{F7CAC9}
\definecolor{RoseQuartz}{HTML}{F5A798}
\definecolor{Serenity}{HTML}{92A8D1}
\definecolor{OrangeRed}{rgb}{1.0, 0.27, 0.0}
\definecolor{Red}{rgb}{1.0, 0.0, 0.0}
\definecolor{Turquoise}{HTML}{0F4C81}
\usepackage{xparse}
\usepackage{soul}

\usepackage{amssymb}
\usepackage{array}
\newcolumntype{L}{>{\centering\arraybackslash}m{4cm}}
\newcolumntype{M}{>{\centering\arraybackslash}m{3cm}}
\newcolumntype{S}{>{\centering\arraybackslash}m{2cm}}
\newcolumntype{P}{>{\arraybackslash}m{12cm}}
\newcolumntype{Q}{>{\arraybackslash}m{6cm}}
\usepackage{xspace}
\usepackage{xcolor}
\usepackage{ulem}
\usepackage{mdframed}
\usepackage{fontawesome}
\definecolor{ao}{rgb}{0.0, 0.5, 0.0}
\definecolor{forestgreen}{rgb}{0.13, 0.55, 0.13}
\usepackage{xspace}

\newtcolorbox{promptbox}[1]{colback=tablerow1!5!white,
colframe=tablerow1!75!black,fonttitle=\bfseries,
title={#1}, left=2mm, right=2mm}
\title{Debate-to-Write: A Persona-Driven Multi-Agent Framework for Diverse Argument Generation}

\author{Zhe Hu$^{1}$,  Hou Pong Chan$^{2}$,  Jing Li$^{1,3}$, Yu Yin$^{4}$
\\
  $^{1}$Department of Computing, The Hong Kong Polytechnic University\\
  $^{2}$DAMO Academy, Alibaba Group\\
  $^{3}$Research Centre for Data Science \& Artificial Intelligence \\
  $^{4}$Department of Computer and Data Sciences, Case Western Reserve University
 \\
  $^{1}${\tt zhe-derek.hu@connect.polyu.hk, jing-amelia.li@polyu.edu.hk} \\
  $^{2}${\tt houpong.chan@alibaba-inc.com}\quad
  $^{4}${\tt yxy1421@case.edu}
  }

\begin{document}
\maketitle

\begin{abstract}
Writing arguments is a challenging task for both humans and machines.
It entails incorporating high-level beliefs from various perspectives on the topic, along with deliberate reasoning and planning to construct a coherent narrative. Current language models often generate outputs autoregressively, lacking explicit integration of these underlying controls, resulting in limited output diversity and coherence.
In this work, we propose a persona-based multi-agent framework for argument writing. Inspired by the human debate, we first assign each agent a persona representing its high-level beliefs from a unique perspective, and then design an agent interaction process so that the agents can collaboratively debate and discuss the idea to form an overall plan for argument writing. Such debate process enables fluid and nonlinear development of ideas. We evaluate our framework on argumentative essay writing. The results show that our framework generates more diverse and persuasive arguments by both automatic and human evaluations.
~\footnote{We release our code and data at:~\url{https://github.com/Derekkk/LLM4ArgGen}.} 

\end{abstract}

\section{Introduction and Related Work}

One of the most common formats of opinion-based communication is argumentation, where users present their viewpoints and attempt to persuade others to adopt their stance on various topics. 
Writing argumentative essays on controversial topics presents significant challenges in natural language processing~\cite{hua-wang-2018-neural,wang2023argument,hua-etal-2019-argument-generation}. The complexity of this task stems from several requirements: Firstly, it necessitates social understanding capabilities for a profound comprehension of the topic and the inclusion of varied, pertinent viewpoints to bolster the argument's persuasiveness. Secondly, it demands strong logical reasoning and strategic text planning to create a coherent overarching structure, which integrates different viewpoints into a well-organized discourse. Lastly, fundamental writing skills are crucial for effectively transforming the plans into surface text.

Large language models (LLMs) have demonstrated impressive outcomes~\cite{ouyang2022training,touvron2023llama,achiam2023gpt}. Yet, they still face challenges when tasked with argument generation~\cite{hu-etal-2024-americano-argument,he-etal-2024-decomposing}. 
One significant limitation is that they struggle to provide diverse and rich perspectives, particularly when generating subjective content that includes multiple viewpoints~\cite{muscato-etal-2024-overview,hayati2023far}. This limitation stems from the fact that LLMs are trained to model averages and may overlook the nuance and in-group variation of perspectives~\cite{sorensen2024value}. However, the ability to present diverse perspectives is crucial for crafting persuasive arguments that resonate with a broad audience.~\footnote{Instead of the inherent diversity within a single argument, we focus on the ability to generate multiple distinct outputs from the same input, to various groups or audiences.} An ideal system should be capable of tailoring its outputs to diverse sociodemographic groups, ensuring that it remains inclusive and avoids bias toward any dominant or singular viewpoint~\cite{padmakumar2024does}.


Additionally, current LLMs often generate text autoregressively without explicit planning~\cite{bubeck2023sparks,wang2022language}, contrasting with human writing that typically involves extensive planning to establish a coherent high-level logic flow~\cite{flower1981cognitive,hu-etal-2022-mocha}. Recent efforts address this by decomposing the generation into content planning and surface writing~\cite{yang2022re3,zhou2023recurrentgpt}. These methods have proven effective for narrative texts, such as stories. However, planning for argumentative texts is inherently more complex. Unlike story outlines that unfold step-by-step sequentially, arguments require weaving together multiple perspectives to form a cohesive and persuasive logic flow. This demands nonlinear thinking~\cite{tong2023eliminating} to build a robust logical structure, integrate diverse viewpoints effectively, and anticipate and address potential counterarguments. Such complexity necessitates a more sophisticated approach to planning in LLMs for argumentation.



\begin{figure*}[t]
    \centering
    \includegraphics[scale=0.5]{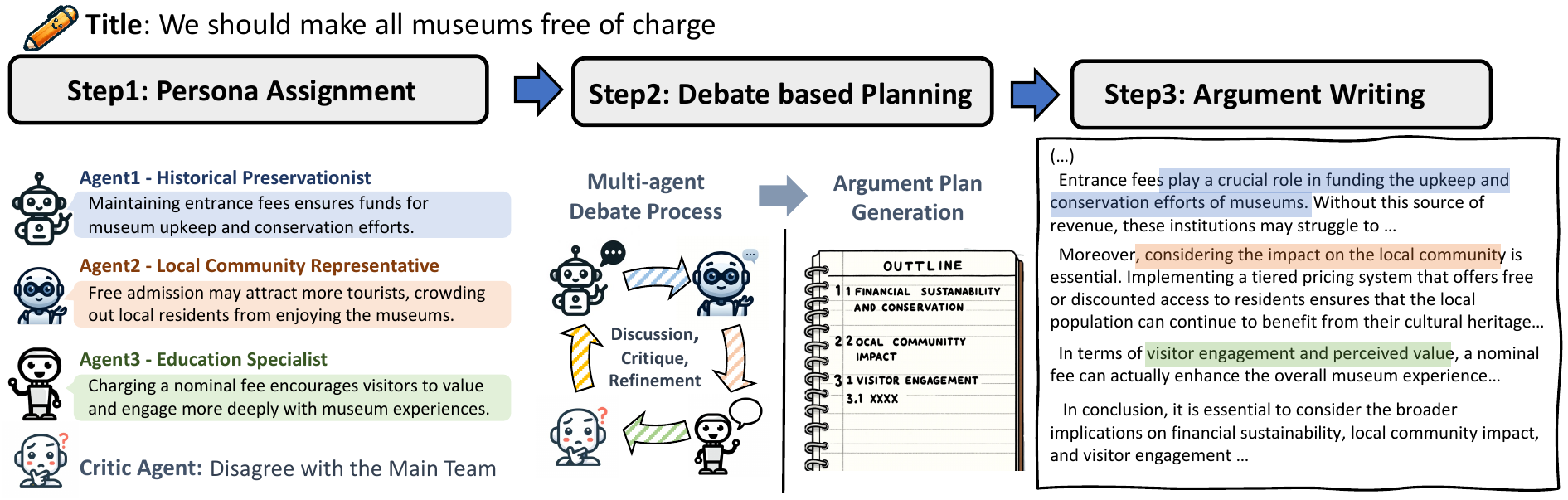}
    \vspace{-3mm}
    \captionof{figure}{The overview of our framework. Given an input topic, our framework first assigns distinct personas to each agent, each representing a unique perspective relevant to the topic. The agents then engage in discussions and debates to refine their ideas and develop a high-level plan. Finally, an argument writing module transforms this plan into a surface argumentative essay. A complete sample output with intermediate results for each step is shown in Figure~\ref{fig:additonal_samples2}.
    }
    \vspace{-3mm}
    \label{fig:overall}
\end{figure*}

In this paper, we propose a persona-based multi-agent framework built upon LLMs for writing argumentative essays that are \textit{perspective-diverse} and \textit{logically coherent}. Recent work shows that assigning personas to LLMs can enhance the performance towards specific perspective and believable human behavior~\cite{jiang2023personallm,xu2023expertprompting}. To enhance perspective diversity, our framework employs multiple agents, each endowed with a distinct persona, representing a unique viewpoint relevant to the input topic. This multi-persona collaboration brings unique perspectives and expertise to the table, thus crafting a more compelling and persuasive argument~\cite{johnson2006logical}. 

Inspired by previous work utilizing multi-agent debate to improve LLMs' performance~\cite{wang2023unleashing,du2023improving,DBLP:journals/corr/agent-faithful-explanation24}, we model text planning as a debate process among the agents. 
During the debate, agents engage in dialogue, respond to critiques, and progressively refine their ideas. 
This collaboration not only fosters creativity and critical thinking but also aids in self-revision and self-critic. 
The discussions are then distilled into an argument plan that offers diverse viewpoints and maintains logical coherence. Unlike previous planning methods that \textit{sequentially} outline content~\cite{hu-etal-2022-planet,goldfarb-tarrant-etal-2020-content,yang2022re3}, our debate-driven planning allows \textit{fluid and nonlinear development} of ideas, where agents can dynamically shift between proposals, revisit earlier concepts, and organically evolve the discussion. 


Furthermore, current evaluation metrics for content diversity primarily measure lexical or semantic diversity, making them insufficient for assessing perspective diversity in long-form discourse. 
To evaluate the diversity of perspectives a model can provide in generating arguments, we propose a novel automatic metric on perspective diversity. This metric works by extracting key ideas, assessing the uniqueness of each perspective, and aggregating these scores to determine the overall diversity of perspectives. This approach effectively measures the range of viewpoints the model incorporates in its argument generation.

We conduct experiments on argumentative essay writing using topics from the idebate and reddit/CMV portals, encompassing a wide range of domains. Both automatic and human evaluations indicate that our method produces outputs that are more diverse and coherent compared to those generated by baselines. Our key contributions are: 

\begin{itemize}[wide,nolistsep]
\item We propose a persona-based multi-agent approach to ensure diverse perspectives in argument generation;
\item We develop a debate-driven planning that allows fluid and nonlinear development of ideas;
\item We design a novel metric for evaluating perspective diversity in long-form output.

\end{itemize}


\section{Method}

Given an input proposition ($x$) on a topic, our multi-agent framework generates an argument ($y$) with the following steps: (1) persona assignment, which creates and assigns an underlying persona to each agent; (2) debate-based planning, where agents collaboratively engage in debate and discussion to form a high-level plan; (3) argument writing that transforms the developed plan into a surface argument. 
All modules are implemented using LLMs with prompting, which eliminates the need for additional model training efforts. The overall framework is shown in Figure~\ref{fig:overall}.

\subsection{Persona Assignment} 
Faced with a proposition on a controversial topic, people often form their opinions based on their underlying beliefs. This module generates and assigns a unique persona to each agent, representing their core beliefs. These personas serve as hidden variables that influence the agents' contributions during subsequent debate and writing tasks.

\smallskip
\noindent\textbf{Persona Pool Creation.}
We instruct LLMs to create a pool of personas, each embodying a distinct viewpoint relevant to the topic. We formalize a persona with a brief description and a claim on the topic, 
as illustrated in Figure~\ref{fig:overall}. To ensure fairness and inclusivity, the model is directed to create personas representing a diverse range of communities and perspectives.

\smallskip
\noindent\textbf{Persona Selection.}
After creating the persona pool, LLMs are prompted to select a combination of $N$ personas from the pool and assign them to each participant, where $N$ represents the number of participants. The model is guided to provide an explanation for each persona selection, ensuring that the chosen personas collectively contribute to a robust collaborative effort. We set $N$ as 3 in our work.

\subsection{Multi-agent Debate for Text Planning}
Recent studies have highlighted the effectiveness of improving LLM performance with
multi-agent collaboration~\cite{li2023prd,wang2023unleashing,du2023improving,liang2023encouraging}. We introduce a persona-based multi-agent debate for text planning, with each agent implemented as an LLM instance.

In this framework, $N$ agents, along with a \textit{critic agent}~\cite{du2023improving}, engage in structured debates to collaboratively develop a plan that outlines the high-level logical flow. The critic agent adopts an opposing viewpoint, actively identifying and challenging weaknesses in the proposals put forth by the other agents. This approach encourages anticipation of opposing perspectives and promotes the formulation of more robust, well-rounded plans through iterative discussion and rebuttal.

After the agent initialization, the agents start a debate and express their opinions iteratively. This discourse continues multiple rounds until reaching a consensus. Subsequently, the model synthesizes a final argument plan based on the discussion, representing the high-level logical flow. Our debate-driven planning mirrors real-time discussions, wherein ideas evolve, face challenges, and undergo refinement in a nonlinear manner.



\subsection{Argument Writing}

The argument writing module then transforms the plan into a final argument. By employing the plan as high-level guidance, this module 
generates arguments in a controllable manner
to ensure the output coherence. Our framework promotes thoughtful deliberation in the writing process by decomposing the text planning stage from end-to-end generation, enabling more polished and structured arguments.

\section{Experiment Setup}


\subsection{Tasks}
We evaluate our framework on argument essay writing~\cite{bao-etal-2022-aeg}. We collect propositions from \url{idebate.net} and reddit/CMV on various domains such as culture, politics, and education. Each proposition represents a controversial topic, like \textit{``We should make all museums free of charge.''} 
A model needs to generate a counter-argumentative essay to refute the proposition. We randomly sample 64 inputs for evaluations. The full list of inputs are shown in Figure~\ref{fig:input_query}.

\subsection{Model and Baselines}
We implement all modules by prompting an LLM. For baselines, we include: (1) Directly prompting an LLM ({E2E}) to write an argument essay; (2) Chain-of-Thought Prompting for content planning ({CoT-Plan}), where the model first generates a plan sequentially and then produces the argument~\cite{wei2022chain}; 
(3) \textsc{Americano}: decomposed argument generation with discourse-driven planning~\cite{hu-etal-2024-americano-argument}. We also include our model variants without persona assignment, as ablations.
We utilize ChatGPT as the backbone LLM, and use the \textit{gpt-3.5-turbo-0301}\footnote{\url{https://platform.openai.com/docs/models}} version. During inference, we set the temperature parameter as 1.0. More implementation details are in Appendix~\ref{sec:appendix_exp_details}.

\subsection{Evaluations}
\smallskip
\noindent\textbf{Quality.} We employ both automatic and human evaluations. For \underline{automatic evaluations}, as current semantic-based metrics such as BLEU and ROUGE do not align well with human judgement~\cite{celikyilmaz2020evaluation,hu-etal-2024-americano-argument}, we follow previous work and utilize GPT-based methods, including for \textit{human preferences}~\cite{NEURIPS2023_91f18a12} and \textit{relevance}~\cite{chia2023instructeval} of generated arguments.
For \underline{human evaluation}, we assess the \textit{persuasion} and \textit{overall quality} of generated arguments. More details are in Appendix~\ref{sec:eval_details}.

\smallskip
\noindent\textbf{Diversity.} We prompt a model to generate 7 arguments for each input and compute both \textit{semantic} and \textit{perspective} diversity of the outputs.
For \textit{semantic diversity}, we use self-BLEU~\cite{zhu2018texygen} to measure diversity among multiple generations for an input. As BLEU score only captures word overlap, we also propose a self-Emb metric where we use embedding similarity to replace the BLEU score.

To evaluate \textit{perspective diversity}, 
we introduce a novel metric that quantifies how many distinct perspectives the model can generate when constructing multiple arguments. 
In particular, we first extract the key perspective points from arguments by ChatGPT, and then calculate the semantic overlap with perspectives of other outputs generated from the same input. This approach assesses the model’s ability to produce multiple outputs with distinct viewpoints. Details are in Appendix~\ref{sec:auto_eval_details}.


\section{Result and Analysis}
\begin{table}[t]
\fontsize{8}{11}\selectfont
 \setlength{\tabcolsep}{1.5mm}
  \centering
    \begin{tabular}{l cc c c c}
        \toprule 
  
& \multicolumn{2}{c}{\textbf{Quality Eval. (↑)}}
& \multicolumn{3}{c}{\textbf{Diversity Eval. (↓)}}   \\ \cmidrule(l){2-3} \cmidrule(l){4-6}  
        {\bf Method}  & {\bf Rel. } & {\bf Pref.} & {\bf S-BLEU} & {\bf S-Emb} & {\bf Pers.} \\
        \midrule
        \text{E2E}  & 3.56 & 3.63 & 24.54 & 87.66 & 73.77\\
        \text{CoT-Plan} & 3.59&  3.48 & 20.54 & 86.54 & 73.37 \\
        \textsc{Americano} &  3.75 & \textbf{4.14}  & 22.93 &  85.59 & 72.85 \\
        \text{Ours}& \textbf{3.89} &3.91  &   \textbf{18.61} &\textbf{84.91} & \textbf{70.71}\\
        \quad w/o persona &3.81 &3.75   & 19.70 &  85.60 & 71.71 \\
        \bottomrule
    \end{tabular}
    \vspace{2mm}
    \caption{
    Automatic results. For quality, we evaluate output relevance (Rel.) and preferences (Pref.). For diversity, we report self-BLEU (S-BLEU), self-Emb (S-Emb) and Perspective Diversity (Pers.).
  }
  \label{tab:automatic_results}
  \vspace{-6mm}
\end{table}

\subsection{Main Results}
The automatic results are shown in Table~\ref{tab:automatic_results}. For quality evaluation, our model achieves the highest scores for output relevance and the second highest score for human preferences, demonstrating its effectiveness. Moreover, Our model variant without the persona module outperforms directly prompted (E2E) and sequential planning (CoT-Plan) baselines, underscoring the efficacy of leveraging multi-agent debate for text planning to enhance model performance in argumentation. However, it underperforms ours, indicating the advantage of the persona module for improving argument quality.

For diversity, our model significantly surpasses all baselines, producing outputs with both semantic diversity and rich perspectives. Conversely, LLM-E2E generates the least diverse outputs in terms of perspectives. This proves the effectiveness of persona assignment to enable the model to encompass a broader spectrum of viewpoints.

\begin{table}[t]
\fontsize{9}{11}\selectfont
 \setlength{\tabcolsep}{1.5mm}
  \centering
    \begin{tabular}{lcc}
        \toprule 
        {\bf Model}  & {\bf Persuasiveness} & {\bf Overall} \\
        \midrule
        E2E  &  2.05 / 41.4\%  & 1.99 / 36.8 \%\\
        CoT-Plan & 2.21 / 48.3\%& 2.13 / 39.1 \% \\
        Ours & \textbf{2.31} / \textbf{51.7 \%} & \textbf{2.47} / \textbf{66.7 \%} \\
        \bottomrule
    \end{tabular}
    \vspace{2mm}
    \caption{
    Human evaluation results. The first position is the score, and the second position is the percentage of results ranked first (ties are allowed).
  }
  \label{tab:human_eval_res}
  \vspace{-4mm}
\end{table}

\begin{figure}[t]
    \centering
    \includegraphics[scale=0.49]{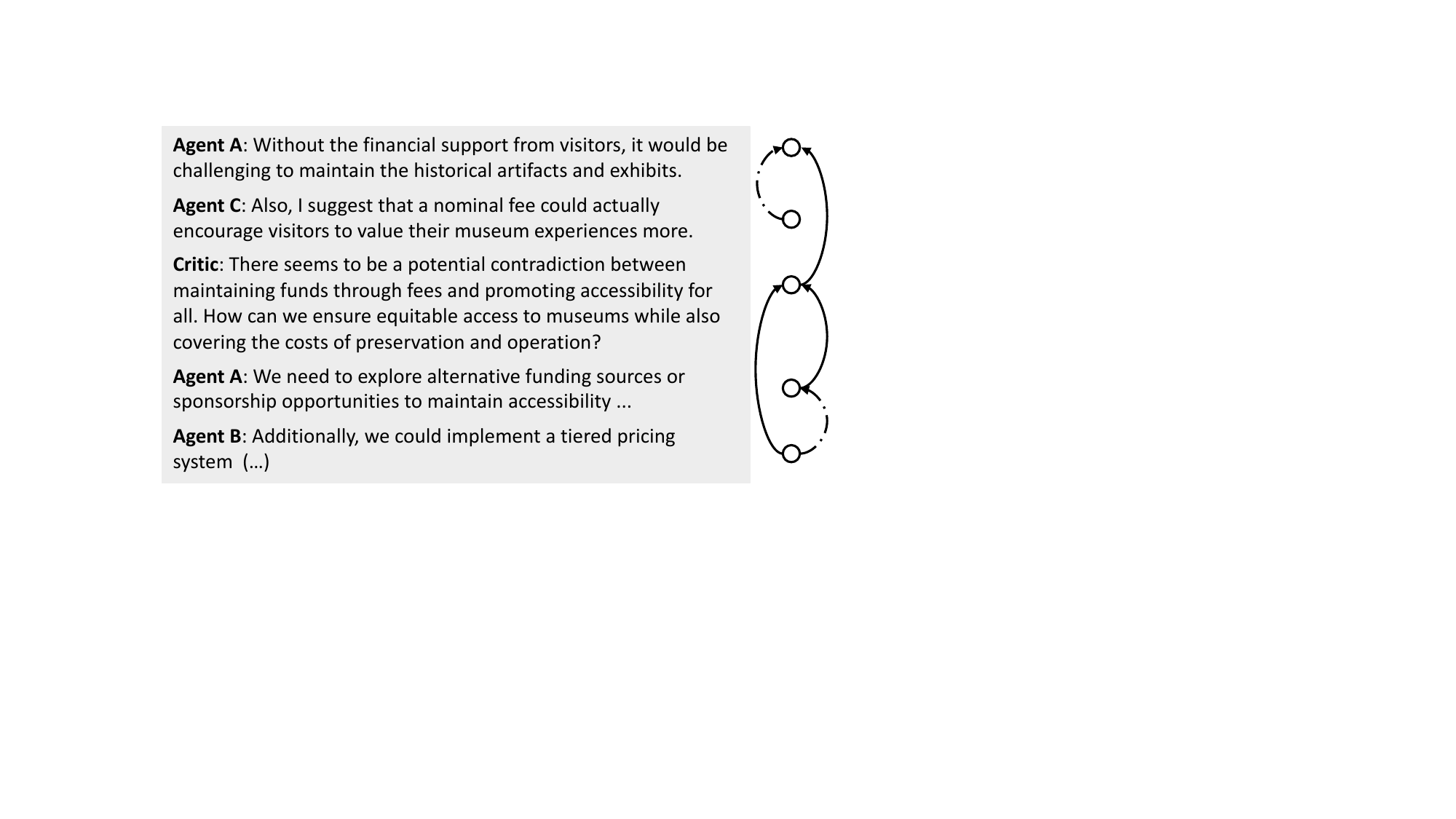}
    \vspace{-6mm}
    \captionof{figure}{Snippet of the debate among agents for example in Figure~\ref{fig:overall}. The right structure shows the logical flow, where solid arrow is oppose relation and dashed arrow is support.
    }
    \vspace{-4mm}
    \label{fig:sample_discussion}
\end{figure}

\begin{figure*}[t]
    \bgroup
    \def\arraystretch{1.5}
	\fontsize{7}{7}\selectfont
     \hspace{-1mm}
	\setlength{\tabcolsep}{0.8mm}
	\centering
	\begin{tabular}{p{155mm}}
	\midrule	
	\textbf{Topic}: We should make all museums free of charge \\
	\midrule	
\textbf{Agents of main team 1}\\
- Agent A - A museum employee: Making museums free would lead to budget cuts that could prevent museums from providing the quality of exhibits and educational programming.\\
- Agent B - An art collector: A free admission policy would lead to an influx of visitors who are not genuinely interested in the art, leading to more congestion, less space and consequently less enjoyment for art lovers. \\
- Agent C - A taxpayer: Free admission to museums would result in increased taxes, which would not only harm low-income individuals but would also impose an unnecessary burden on working and middle-class families who are already struggling to meet ends.
\\ 
	\midrule	
\textbf{Agents of main team 2}\\
- Agent A - A historian who values the preservation of artifacts: If museums are free, they may not generate enough revenue to properly maintain and protect historic artifacts and valuable works of art.\\
- Agent B - A community organizer advocating for equal access to cultural institutions: Charging admission fees to museums disproportionately affects low-income communities and excludes them from engaging in cultural experiences. \\
- Agent C - A museum curator trying to balance the budget: Without admission fees, museums may struggle to generate enough revenue to cover operational costs, resulting in reduced services and potentially closing the museum altogether.
\\ 
\midrule
\textbf{Agents of main team 3}\\
- Agent A - A museum administrator: Making museums free would result in the loss of crucial funding needed to maintain and improve exhibits.\\
- Agent B - A cultural studies scholar: Charging for museum admission is actually beneficial as it allows people to value the works more and fosters cultural appreciation.\\
- Agent C - A museum donor: Eliminating museum admission fees would deter future charitable donations and hamper the ability of museums to acquire new exhibits and fund construction projects.\\
    \bottomrule
    
	\end{tabular}
    \vspace{-2mm}
    \caption{Different Agent persona assignment for the same topic.} 
\label{fig:multiple_assignment}
\egroup
\vspace{-2mm}
\end{figure*}

\subsection{Human Evaluations}
Due to the limitation of automatic evaluations, we also conduct human evaluations. 
We compare our model with E2E and CoT-Plan considering the substantial effort for evaluating arguments. We randomly sample 30 inputs, and ask three human judges to evaluate the models outputs on aspects of \textit{persuasiveness} and \textit{overall quality}. Details are in Appendix~\ref{sec:appendix_human_eval}. 

As shown in Table~\ref{tab:human_eval_res}, human judges consistently rate our model outputs better than the baselines in both aspects. Particularly, our model generates outputs that cover a broader range of perspectives, thereby enhancing the overall persuasiveness of the argument. Moreover, our model is more frequently ranked as the top choice, further demonstrating its effectiveness in generating persuasive and high-quality argumentative essays.

\subsection{Sample Analysis}
\smallskip\noindent\textbf{Analysis on Debate Process.}
In Figure~\ref{fig:sample_discussion}, we show a snippet of debate process with logic structure for the input \textit{``We should make all museums free of charge''} as in Figure~\ref{fig:overall}. 
The right structure shows the logical flow in a nonlinear manner, where the agents not only progressively discuss the idea but also revisit and revise earlier points to address the critics. Such debate process mimics a real-time discussion where ideas are constantly evaluated, challenged, and refine. By fostering an environment of continuous dialogue and reflection with nonlinear thinking, the internal multi-agent debate creates a more flexible and comprehensive planning process.

\smallskip\noindent\textbf{Analysis on Persona Assignment.}
We show persona assignments for the same topic generated from multiple times, as in Figure~\ref{fig:multiple_assignment}. The persona and claim represent the underlying themes of each perspective. As we can see, our model generates distinct viewpoints for the same topic, enhancing diversity in perspectives. Such sampling explicitly encourages the model to consider different perspectives and viewpoints on the topic, thereby leading to a better diversity of multiple generations. We provide more sample outputs and discussions in Appendix~\ref{sec:appendix_samples}.

\section{Conclusion}
In this study, we introduce a multi-agent debate framework with persona assignment for each agent to enrich perspective diversity and enhance persuasiveness in argument generation. Our debate-driven planning fosters fluid and nonlinear development of ideas for text planning, resulting in more robust and coherent argument plans. Experimental results across diverse topics demonstrate that our framework yields more diverse and superior arguments. Future work includes devising supervised fine-tuning~\cite{DBLP:journals/corr/cmat24} or reinforcement learning techniques~\cite{ouyang2022training,DBLP:journals/taslp/HuCCLXSW23,DBLP:journals/tacl/ChanWK21} to further enhance the collaborative writing capability of LLMs. 

\section*{Acknowledgments}
This work is supported by a grant from the Research Grants Council of the Hong Kong Special Administrative Region, China (Project No. PolyU/25200821), the NSFC Young Scientists Fund (Project No. 62006203), the Innovation and Technology Fund (Project No. PRP/047/22FX), and PolyU Internal Fund from RC-DSAI (Project No. 1-CE1E). Additionally, we thank all the reviewers for their useful feedback. 

\section*{Limitations}
Our work has several limitations that could be addressed in future studies. Firstly, effective argumentative essays often rely on supporting evidence to bolster claims. Humans typically seek out relevant knowledge or evidence to augment the persuasiveness of their arguments. Therefore, our framework could benefit from the integration of a knowledge retrieval module to incorporate external evidence. 


In addition, we acknowledge the relative small test set due to the reason that argument essay writing is a long-form, open-ended generation task, where the complexity and length of outputs make both automatic and human evaluations more challenging and resource-intensive. However, our dataset is carefully curated to cover a wide range of themes and topics in argumentation, which we believe provides a robust basis for evaluating our framework's performance. 

Finally, in our experiments, GPT-3.5 serves as the base model. However, other LLMs, including smaller models (e.g., 7B and 13B), can also be integrated to further demonstrate the effectiveness of our framework. Exploring the performance of these models will be a focus of future work.



\section*{Ethics Statement}
Acknowledging the reliance of our framework on large language models, we recognize the possibility of generating fabricated and potentially harmful content due to inherent biases in the pre-training data drawn from heterogeneous web corpora for LLMs. Given the inability to fully control the language model generation process, there exists a risk of unintended biases persisting in the generated outputs. We strongly urge users to meticulously evaluate the ethical implications of the generated content and exercise prudence when employing the system in real-world contexts.

\bibliography{custom}

\appendix

\section{Experimental Details}
\label{sec:appendix_exp_details}
\subsection{Dataset}
In this work, we study zero-shot argumentative essay writing leveraging the large language models, and we select topics from \url{idebate.net} and reddict/CMV~\footnote{\url{https://www.reddit.com/r/changemyview/}}. Each topic is  a controversial proposition, such as \textit{``We should make all museums free of charge.''} We select 64 inputs covering different domain, and ensure they do not contain offensive contents. The full list of inputs are shown in Figure~\ref{fig:input_query}.
The model is asked to write a counter-argumentative essay to refute the proposition. 

\subsection{Model Details}
We set the number of agents for the main team as 3 in our experiments.
We utilize ChatGPT as the backbone LLM, and use the \textit{gpt-3.5-turbo-0301}\footnote{\url{https://platform.openai.com/docs/models}} version. During inference, we set the temperature parameter as 1.0. During the planning process, we define the plan as a high-level outline that contains several main points, where each point can be supported by several sub-points. We also allow an optional acknowledgment point. The specific prompts we leveraged are presented from Figure~\ref{fig:prompt_persona_assignment} to Figure~\ref{fig:prompt_surface_gen}.

\smallskip
\noindent\textbf{Baselines.} (1) For E2E, we directly prompt an LLM to generate the output without explicit text planning. (2) For CoT-Plan, we first prompt an LLM to write a high-level plan, and the generate output based on the topic and plan. Similar to our model, we define plan as the same structure. This baseline is similar to the chain-of-thought prompting where the model first think about the high-level contents by generating the plan and then producing the final output. 
(3) \textsc{Americano}~\cite{hu-etal-2024-americano-argument} is an argument generation framework that decomposes the generation based on argumentative discourse structures.

\section{Evaluation Details}
\label{sec:eval_details}
\subsection{Automatic Evaluation Details}
\label{sec:auto_eval_details}
For GPT-based evaluation, we leverage the GPT4 model with the \textit{``gpt-4o-2024-05-13''} variant. The prompt used for evaluating \textbf{relevance}~\cite{chia2023instructeval} and \textbf{human preference}~\cite{NEURIPS2023_91f18a12} are adopted from the original papers.
For \textbf{diversity} evaluation, to compute the embedding diversity, we apply \textit{"text-embedding-3-small"} model from OpenAI API to transform each output to an embedding, and then compute their cosine similarity. 

\smallskip\noindent\textbf{Semantic Diversity}.
Besides self-BLEU, we design a self-Emb method, where we use cosine similarity between two output embeddings to replace the BLEU score. we apply \textit{"text-embedding-3-small"} model from OpenAI API to transform each output argument to an embedding.

\smallskip\noindent\textbf{Perspective Diversity}.
We introduce a novel metric that quantifies how many distinct perspectives the model can generate when constructing multiple arguments. 
To calculate the \textit{perspective diversity} score, for each input, a model generates $M$ arguments $\{y_1,...,y_M\}$. For a generated argument $y_m$, we first extract its main perspective points $O_m=\{o_{m1},...,o_{mn}\}$ by prompting ChatGPT with the following prompts:

\begin{promptbox}{Perspective Extraction Prompt}
Given an argumentative essay: 

[ARGUMENT]

Summarize the opinions of the essay with at most 5 points, with each line as one point. Use as fewest points as you can to make the output abstract and concise.
\end{promptbox}

Then for each perspective point $o_{mi}$, we compute its embedding similarity with the perspective points from all other $M-1$ arguments generated with the same input by computing the embedding similarity, and take the maximum similarity score ($s_{mi}$). The perspective diversity score of $y_m$ is then computed as $s_m = \frac{1}{n}\sum_{1}^{n}s_{mi}$. The overall diversity score of the sample is the average of all arguments: $\frac{1}{M}\sum_{1}^{M}s_{m}$.  A lower score indicates better perspective diversity achieved.

\begin{figure*}[t]
    \centering
    \includegraphics[scale=0.2]{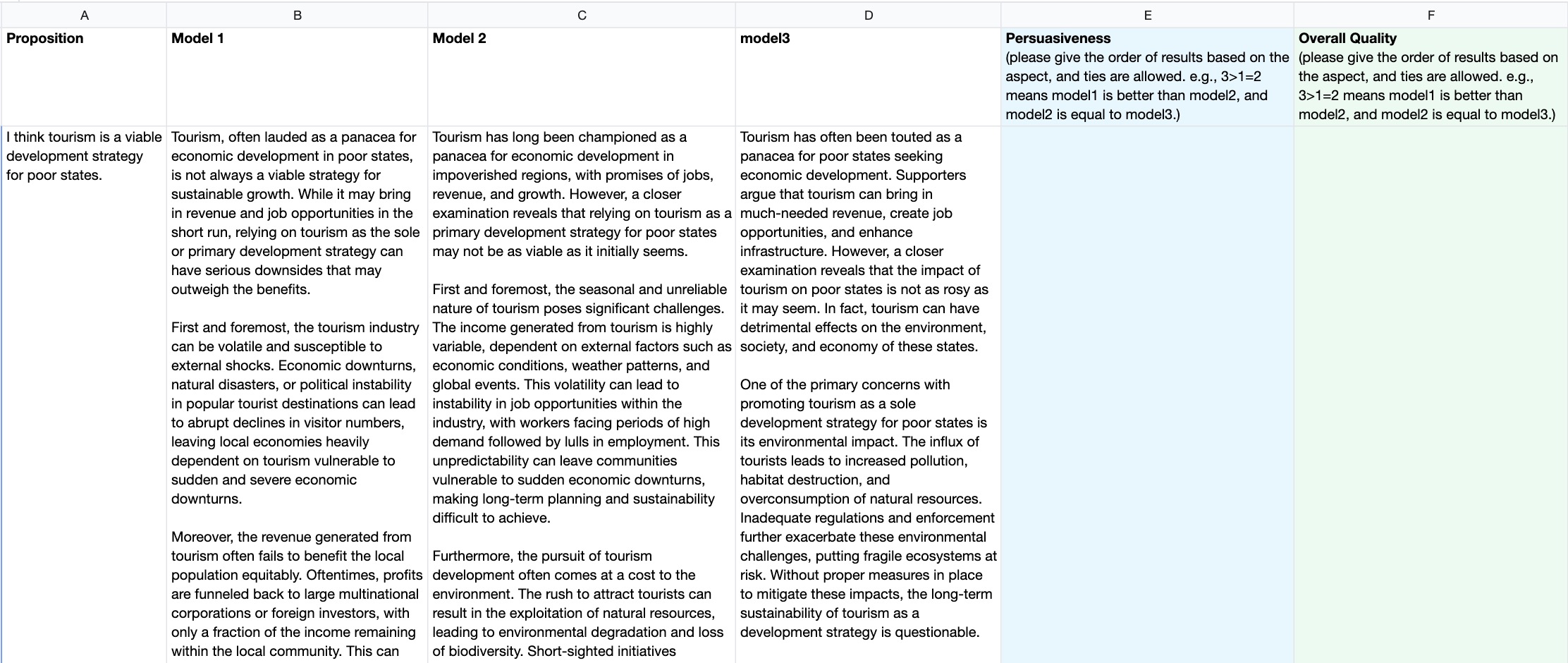}
    \vspace{-2mm}
    \captionof{figure}{The interface of human evaluations.
    }
    \vspace{-2mm}
    \label{fig:interface}
\end{figure*}

\subsection{Human Evaluation}
\label{sec:appendix_human_eval}
For human evaluation, we hire three human judges to evaluate the model outputs on persuasiveness and overall preference. We ensure all judges are proficient English speakers with at least a Bachelor degree. All judges are graduate students based in US, and we pay them \$12 per hour.
We randomly select 30 inputs, and for each input we present the outputs of different models anonymously. 
We ask the human judges to rank the outputs based on each evaluation aspect, and ties are allowed. The evaluation interface is shown in Figure~\ref{fig:interface}.

Specifically, for \textit{persuasiveness aspect}, we ask the human judges to determine: whether the essay effectively challenges the initial proposition by providing convincing viewpoints from various perspectives with coherent logic; whether it is likely to persuade you to reconsider your initial position. For \textit{overall quality}, we ask the human judge to evaluate on its overall quality and writing and then rank the outputs.

After evaluation, we convert the ranking results into scores by subtracting its ranking position from the total number of outputs. For example, a ranking order of \textit{$model_1$ > $model_2$ > $model_3$} will lead to the score of 3 for $model_1$, 2 for $model_2$, and 1 for $model_3$. We then report the average scores of each model, as in Table~\ref{tab:human_eval_res}. This shows the relative performance of each model. We also compute the percentage of times a model output is considered as the best (i.e., ranked first).

\section{Sample Outputs and Additional Analysis}
\label{sec:appendix_samples}

We present several comprehensive sample outputs generated by our models from Figure~\ref{fig:additonal_samples1} to Figure~\ref{fig:additonal_samples3}. 
These examples showcase each step of our process, demonstrating how our framework effectively models a debate and discussion among agents. As we can see, the agents develop and refine ideas in alignment with their assigned personas, while the critic agent plays a crucial role by challenging proposals, prompting revisions, and addressing potential rebuttals. After the debate process, the model constructs high-level logical plans derived from the discussion, guiding the subsequent formulation of final arguments.

Despite these strengths, our manual inspection of the outputs reveals a tendency for the arguments to rely heavily on reasoning without sufficient evidential support. For instance, the argument generated in Figure~\ref{fig:additonal_samples1} puts forward "\textit{Financial Sustainability and Conservation}" as a key claim but lacks concrete examples or research to substantiate this position. Persuasive arguments need to be be grounded in factual evidence or expert opinions to effectively reinforce their claims. To enhance the overall persuasiveness of our models, future work should investigate the integration of explicit knowledge retrieval to provide stronger evidential backing.

\begin{figure*}[t]
    \bgroup
    \def\arraystretch{1.8}
	\fontsize{8}{8}\selectfont
     \hspace{-1mm}
	\setlength{\tabcolsep}{0.8mm}
	\centering
	\begin{tabular}{p{155mm}}
	\toprule
	{\textbf{Argument Plan}}:\\
\circled{1} Acknowledgment:\\
\quad\quad- Recognize the value of free admission in promoting accessibility and attracting tourists.\\
\circled{2} Financial Sustainability and Conservation\\
\quad\quad- Entrance fees are crucial for funding museum upkeep and conservation efforts.\\
\quad\quad- Alternative funding sources and sponsorships can supplement revenue without hindering accessibility.\\
\circled{3} Local Community Impact\\
\quad\quad- Implement a tiered pricing system to ensure locals have free or discounted access.\\
\circled{4} Visitor Engagement and Value\\
\quad\quad- A nominal fee can encourage visitors to engage more deeply with museum experiences.\\
\quad\quad- Thoughtful pricing strategies can enhance the overall value perception for visitors.\\
\midrule
{\textbf{Argumentative Output}}:\\
While the notion of providing free admission to all museums may seem appealing at first glance, a deeper examination reveals that this proposition might not be as beneficial as it appears. Undoubtedly, there are advantages to offering free access, such as promoting inclusivity and attracting tourists. However, the financial sustainability and conservation of these cultural institutions must also be taken into account.\\

Entrance fees play a crucial role in funding the upkeep and conservation efforts of museums. Without this source of revenue, these institutions may struggle to maintain their collections, facilities, and educational programs. While it is true that museums can explore alternative funding sources and seek sponsorships, relying solely on these methods may not guarantee the financial stability needed to safeguard and preserve valuable cultural artifacts for future generations.\\

Moreover, considering the impact on the local community is essential. Implementing a tiered pricing system that offers free or discounted access to residents ensures that the local population can continue to benefit from their cultural heritage. By differentiating pricing for tourists, museums can strike a balance where visitors contribute more to support the institutions without excluding those who live in the area. This approach acknowledges the importance of local engagement and the preservation of community ties to these cultural landmarks.\\

In terms of visitor engagement and perceived value, a nominal fee can actually enhance the overall museum experience. When visitors invest financially, even in a small way, they are more likely to engage deeply with the exhibitions and educational resources available. Thoughtful pricing strategies can also create a sense of exclusivity and value, encouraging visitors to appreciate the cultural and historical significance of what they are experiencing.\\

In conclusion, while the idea of making all museums free of charge may seem well-intentioned, it is essential to consider the broader implications on financial sustainability, local community impact, and visitor engagement. By maintaining entrance fees and implementing strategic pricing models, museums can ensure their long-term viability while continuing to provide access to all, thus striking a balance between accessibility, conservation, and cultural value.
 \\
    \bottomrule
	\end{tabular}
    \vspace{-2mm}
\caption{The full generated argument plan and output for input in Figure~\ref{fig:overall} and Figure~\ref{fig:sample_discussion}} 
\label{fig:additonal_samples1}
\egroup
\vspace{-2mm}
\end{figure*}

\begin{figure*}[t]
    \bgroup
    \def\arraystretch{1.8}
	\fontsize{8}{8}\selectfont
     \hspace{-1mm}
	\setlength{\tabcolsep}{0.8mm}
	\centering
	\begin{tabular}{p{155mm}}
	\midrule	
	\textbf{Topic}: We should allow the use of electronic and internet voting in state-organised elections \\
	\midrule	
- Agent A - Elderly voter advocate: Implementing electronic voting may disenfranchise elderly voters who are less familiar with technology and prefer traditional methods. \\
- Agent B - Privacy advocate: Electronic voting raises significant concerns about the privacy and confidentiality of voter information.\\
- Agent C - Skeptical traditionalist: Maintaining the traditional paper ballot system ensures transparency and trust in the electoral process, avoiding potential technological vulnerabilities.
\\ \midrule
\textbf{Debate Process:} \\
Agent A: I believe that implementing electronic voting may disenfranchise elderly voters who are less familiar with technology and prefer traditional methods. This can lead to a decrease in voter participation among the elderly population. \\
Agent B: Additionally, electronic voting raises significant concerns about the privacy and confidentiality of voter information. The security measures in place may not be foolproof, risking the integrity of the electoral process. \\
Agent C: Maintaining the traditional paper ballot system ensures transparency and trust in the electoral process. It avoids potential technological vulnerabilities that could be exploited to manipulate election results. \\

Critic: While the concerns raised are valid, electronic and internet voting can improve accessibility for voters with disabilities and those living far from polling stations. It also has the potential to increase voter turnout among the tech-savvy population. \\

Agent A: I acknowledge the accessibility benefits, but we must prioritize inclusivity for all voter demographics. Introducing electronic voting without addressing the concerns of the elderly population could create a significant divide in voter representation. \\

Agent B: Security should also be a top priority. Even with advancements in technology, the risk of cyber threats and hacking remains a serious issue. We cannot compromise the integrity of elections for the sake of convenience. \\

(...)
\\ \midrule
 
	{\textbf{Argument Plan}}:\\
1. Implementing electronic and internet voting may disenfranchise elderly voters due to technological barriers:\\
\quad\quad- Elderly voters may struggle to adapt to new technology, leading to decreased participation.\\
\quad\quad- Preference for traditional methods should be respected to maintain inclusivity across all age groups.\\
2. Electronic voting raises concerns about privacy and confidentiality:\\
\quad\quad- Security measures may not be completely effective, risking the exposure of voter information.\\
\quad\quad- Safeguarding voter data is crucial to ensure the sanctity of the electoral process.\\
3. Maintaining the traditional paper ballot system ensures transparency and trust:\\
\quad\quad- Paper ballots offer a tangible and easily verifiable voting method.\\
\quad\quad- Avoiding technological vulnerabilities helps uphold the integrity of election outcomes.\\

\midrule
{\textbf{Argumentative Output}}:\\
In the evolving landscape of modern technology, the proposition of implementing electronic and internet voting in state-organised elections has sparked debates among policymakers and citizens. While proponents argue for the convenience and efficiency that digital voting methods can bring, it is essential to critically consider the potential drawbacks and risks associated with such a significant shift in our electoral process.\\

One of the primary concerns with introducing electronic and internet voting is the potential disenfranchisement of elderly voters. This demographic group may face technological barriers that could hinder their ability to participate effectively in the electoral process. The unfamiliarity and complexity of new voting technologies could lead to decreased participation among elderly voters, ultimately undermining the core principle of inclusivity in our democratic system. Respect for the preferences of all age groups, including those who are more comfortable with traditional voting methods, is crucial to ensuring that every eligible voter has an equal opportunity to exercise their democratic right.\\

Furthermore, electronic voting raises legitimate concerns about privacy and confidentiality. The security measures put in place to protect voter information may not always be foolproof, leaving room for potential breaches and hacks that could compromise the integrity of the electoral process. Safeguarding voter data is paramount to upholding the sanctity of elections, and any vulnerabilities in the digital voting system could jeopardize the confidentiality of individual votes and the overall trust in the electoral outcome.\\

In contrast, maintaining the traditional paper ballot system offers a level of transparency and trust that electronic voting may not yet be able to match. Paper ballots provide a tangible and easily verifiable voting method, allowing voters to physically mark their choices and witness the deposit of their ballots into a secure ballot box. By avoiding the inherent technological vulnerabilities associated with electronic and internet voting, the integrity of election outcomes can be better protected, instilling confidence in the legitimacy of the results.\\

In conclusion, while the idea of electronic and internet voting may seem promising in terms of convenience and efficiency, the potential risks and drawbacks cannot be ignored. The disenfranchisement of elderly voters, concerns about privacy and confidentiality, and the importance of maintaining transparency and trust in the electoral process all warrant careful consideration before implementing such a transformative change. As we strive to modernize our electoral system, we must prioritize the security, inclusivity, and integrity of our elections to ensure a fair and reliable democratic process for all citizens.
 \\
    \bottomrule
	\end{tabular}
    \vspace{-2mm}
\caption{A full output of our model generated argument.} 
\label{fig:additonal_samples2}
\egroup
\vspace{-2mm}
\end{figure*}

\begin{figure*}[t]
    \bgroup
    \def\arraystretch{1.8}
	\fontsize{8}{8}\selectfont
     \hspace{-1mm}
	\setlength{\tabcolsep}{0.8mm}
	\centering
	\begin{tabular}{p{155mm}}
	\midrule	
	\textbf{Topic}: No one over the age of 80 should be allowed to serve in government. \\
	\midrule	
- Agent A - Elderly Rights Advocate: Age should not be a determining factor in someone's ability to serve in government as it goes against the principles of equality and non-discrimination.\\
- Agent B - Experienced Politician: Individuals over 80 often bring valuable experience and wisdom to government roles, making them capable and effective leaders. \\
- Agent C - Youth Activist: Limiting government service based on age overlooks the potential of younger generations to bring fresh ideas and perspectives to governance.
\\ \midrule
\textbf{Debate Process:} \\
Agent A: As an Elderly Rights Advocate, I strongly believe that age should not be a barrier to serving in government. Restricting individuals over 80 from government roles goes against principles of equality and non-discrimination.\\

Agent B: I, as an Experienced Politician, argue that older individuals bring valuable experience and wisdom to governance. Excluding them solely based on age can deprive the government of capable leaders.\\

Agent C: I, a Youth Activist, agree that age should not limit one's ability to serve. Younger generations can offer fresh perspectives and innovative ideas to government roles.\\

Critic: I support the proposition that no one over 80 should serve in government. Older age may bring health concerns and potential lack of energy or agility, affecting decision-making and efficiency in governance.\\

Agent A: We acknowledge the concerns about health and agility but believe that individual capabilities should be assessed on merit, not age alone.\\

(...)
\\ \midrule
 {\textbf{Argument Plan}}:\\
1. Age Should Not Determine Government Service\\
\quad\quad- Subpoint: Equality and Non-Discrimination Principles: Every individual should be evaluated based on merit and capabilities, not age.\\
\quad\quad- Subpoint: Capability Is Timeless: Experience and wisdom are valuable assets in governance, regardless of age. \\

2. Embracing Diversity of Perspectives\\
\quad\quad- Subpoint: Fresh Ideas from Younger Generations: Excluding older individuals limits the range of perspectives, hindering innovative governance.\\
\quad\quad- Subpoint: Varied Experiences: Different age groups bring diverse life experiences that enrich decision-making processes.\\

3. Individual Assessment Over Arbitrary Age Limits\\
\quad\quad- Subpoint: Merit-Based Evaluation: Evaluate individuals on relevant criteria like skills, experience, and abilities rather than imposing blanket age restrictions.\\

\midrule
{\textbf{Argumentative Output}}:\\
In a world where age often dictates far too many aspects of life, the proposal that individuals over the age of 80 should be excluded from government service based solely on their age is not only discriminatory but also shortsighted. This notion fails to consider the fundamental principles of equality and non-discrimination that should underpin any fair and just society.\\

Every individual, regardless of age, should be evaluated based on their merit and capabilities. Experience and wisdom, often accumulated over decades, are valuable assets in governance. To dismiss these qualities based on a number is to overlook the timeless nature of capability. As society progresses, we must recognize that age should not be the sole determinant of an individual's ability to contribute to the governance of a nation.\\

Moreover, embracing diversity of perspectives is essential for effective governance. Excluding individuals over 80 limits the range of perspectives and experiences that can enrich decision-making processes. Younger generations may bring fresh ideas, but the wisdom and insight of older individuals are equally valuable in shaping policies that benefit all members of society. Varied experiences, derived from different age groups, contribute to a more holistic and inclusive approach to governance.\\

Instead of imposing arbitrary age limits, we should prioritize individual assessment based on merit. Merit-based evaluation considers relevant criteria such as skills, experience, and abilities, rather than imposing blanket age restrictions that fail to capture the nuanced capabilities of each individual. By evaluating candidates on their qualifications and aptitudes, we ensure that the most competent and capable individuals have the opportunity to contribute to government service.\\

In conclusion, age should not be a barrier to government service. Instead, we should recognize and value the diverse perspectives and experiences that individuals of all ages bring to the table. By embracing merit-based evaluations and rejecting arbitrary age restrictions, we can create a more inclusive and effective governance system that serves the needs of all members of society.
 \\
    \bottomrule
	\end{tabular}
    \vspace{-2mm}
\caption{A full output of our model generated argument.} 
\label{fig:additonal_samples3}
\egroup
\vspace{-2mm}
\end{figure*}

\begin{figure*}[t]
    \def\arraystretch{1.5}
	\fontsize{9}{10}\selectfont
     \hspace{-2mm}
	\setlength{\tabcolsep}{0.8mm}
	\centering
	\begin{tabular}{p{130mm}}
	\toprule
\textbf{Prompt for persona pool creation:}\\
\hdashline
Given a proposition: \#\#input\_proposition

Background: You want to create a pool of 5 to 10 debate agents, who hold the opinions to refute the given proposition from different perspectives. Each agent should present a distinct viewpoint relevant to the proposition.

Task: Assign each agent a unique persona, described in one sentence, along with a corresponding claim that focuses on a specific perspective. Ensure that each agent provides a different viewpoint relevant to the proposition. To promote diversity and fairness, the agents should represent various communities and perspectives.

Please format your persona descriptions as follows, with each line being a json object:

\{"agent\_id": 0, "description": the\_description\_of\_Agent0, "claim": the\_claim\_of\_Agent0\}\\
...
\\
\bottomrule
\textbf{Prompt for persona selection:}\\
\hdashline
Given a proposition: \#\#input\_proposition

You need to build a team of three agents, to work together and collaboratively formulate a persuasive counterargument that refutes the given proposition. 
Now given the following candidates, where each candidate has a unique persona offering a different perspective relevant to the topic at hand. You need to select three agents that you think can together form a strong team to achieve the task. 
You also need to consider the diversity when selecting candidates. For each selection, give the reason why you select the candidate.\\

\#\# Candidate list:\\
\#\#\#candidate\_list\\

Please select three candidates and add a reason. Each line of output should be a json object as follows:\\

\{"agent\_id": 0, "description": the\_description\_of\_Agent0, "claim": the\_claim\_of\_Agent0, "reason": the\_reason\_of\_selection\}\\
...
\\

\bottomrule
	\end{tabular}
\caption{Prompts for persona assignment. 
} 
\label{fig:prompt_persona_assignment}
\vspace{2mm}
\end{figure*}

\begin{figure*}[t]
    \def\arraystretch{1.5}
	\fontsize{9}{10}\selectfont
     \hspace{-2mm}
	\setlength{\tabcolsep}{0.8mm}
	\centering
	\begin{tabular}{p{130mm}}
	\toprule
\textbf{Template for multi-agent debate}\\
\bottomrule
\textbf{Background}\\
\hdashline
\#\# Goal: Modeling a debate process to analyze a given proposition on a controversial topic, and formulate a well-structured counterargument plan to refute the proposition based on the debate discussion.\\

\#\# Additional Guidelines\\
- The discussion should be conducted for multiple rounds until the Main Team members are satisfied with their counterargument plan and Critic is persuaded.\\
- The discussion should provide a rigorous reasoning so that the logic flow is persuasive and coherent.\\
- Plan Quality: The plan should be abstract and concise. It should contain several main points, where each point can be supported by sub-points. There could be an optional acknowledgment point.\\

\bottomrule
\textbf{Main Team Agent}\\
\hdashline
\#\# Participants and Roles\\
A Main Team of three members: Agent A, Agent B, and Agent C\\
- Stance: Oppose the proposition;\\
- Goal: Discuss together to propose a persuasive counterargument plan outlining the overall logical flow to refute the proposition. \\
- Specific Personas and claims of the team members:\\
	- Agent A: persona\_a;\\
	- Agent B: persona\_b; \\
	- Agent C: persona\_c;\\

\bottomrule
\textbf{Critic Agent}\\
\hdashline
\#\# Participants and Roles: A Critic\\
- Stance: Support the proposition;\\
- Goal: You Disagree with the Main Team. Identify and challenge weaknesses in the Main Team's discussion, and debate with the Main team.
\\

\bottomrule
	\end{tabular}
\caption{Prompts for multi-agent debate. 
} 
\label{fig:prompt_debate}
\vspace{2mm}
\end{figure*}

\begin{figure*}[t]
    \def\arraystretch{1.5}
	\fontsize{9}{10}\selectfont
     \hspace{-2mm}
	\setlength{\tabcolsep}{0.8mm}
	\centering
	\begin{tabular}{p{130mm}}
	\toprule
\textbf{Prompt for surface argument writing:}\\
\hdashline
Given a proposition: \{proposition\}\\

Write a persuasive and coherent counterargumentative essay to refute the proposition. 
You should transform the following plan into a coherent essay, which outlines the high-level logical flow of the counterargument. \\
- plan\\
\{plan\}

Note: ensure the essay is coherent and readable. You do not need to include section title from the plan.

- Counterargumentative essay:
\\

\bottomrule
	\end{tabular}
\caption{Prompts for surface argument writing. 
} 
\label{fig:prompt_surface_gen}
\vspace{2mm}
\end{figure*}

\begin{figure*}[t]
    \bgroup
    \def\arraystretch{1.8}
	\fontsize{8}{9}\selectfont
     \hspace{-1mm}
	\setlength{\tabcolsep}{0.8mm}
	\centering
	\begin{tabular}{p{155mm}}
	\toprule
- {"statement": "We should make all museums free of charge", "topic": "Culture"}

- {"statement": "We should return cultural property residing in museums to its place of origin", "topic": "Culture"}

- {"statement": "We should ban beauty contests", "topic": "Culture"}

- {"statement": "I think tourism is a viable development strategy for poor states.", "topic": "Culture"}

- {"statement": "We should restrict advertising aimed at children", "topic": "Culture"}

- {"statement": "I think in a global language", "topic": "Culture"}

- {"statement": "I think science is a threat to humanity", "topic": "Culture"}

- {"statement": "I think that gay couples should not be allowed to adopt kids", "topic": "Culture"}

- {"statement": "We should ban gambling", "topic": "Culture"}

- {"statement": "I think that the feminist movement should seek a ban on pornography", "topic": "Culture"}

- {"statement": "I think the internet encourages democracy", "topic": "Digital Freedoms"}

- {"statement": "I think the internet brings more harm than good", "topic": "Digital Freedoms"}

- {"statement": "We should allow the use of electronic and internet voting in state-organised elections", "topic": "Digital Freedoms"}

- {"statement": "I think that internet access is a human right", "topic": "Digital Freedoms"}

- {"statement": "We should block access to social messaging networks during riots", "topic": "Digital Freedoms"}

- {"statement": "We should not allow companies to collect/sell the personal data of their clients", "topic": "Digital Freedoms"}

- {"statement": "We should ban targeted online advertising on the basis of user profiles and demographics", "topic": "Digital Freedoms"}

- {"statement": "I think politicians have no right to privacy", "topic": "Digital Freedoms"}

- {"statement": "We should ban the use of Digital Rights Management technologies", "topic": "Digital Freedoms"}

- {"statement": "We should block access to websites that deny the Holocaust", "topic": "Digital Freedoms"}

- {"statement": "This house supports the creation of single-race public schools", "topic": "Education"}

- {"statement": "I think that the payment of welfare benefits to parents should be tied to their children", "topic": "Education"}

- {"statement": "I think university education should be free", "topic": "Education"}

- {"statement": "I think that history has no place in the classroom", "topic": "Education"}

- {"statement": "We should make sex education mandatory in schools", "topic": "Education"}

- {"statement": "I think that animals have rights.", "topic": "Environment"}

- {"statement": "This house Believes People Should Not Keep Pets", "topic": "Environment"}

- {"statement": "I think that states should not subsidise the growing of tobacco", "topic": "Environment"}

- {"statement": "I think we're too late on global climate change", "topic": "Environment"}

- {"statement": "This House Belives that wind power should be a primary focus of future energy supply.", "topic": "Environment"}

- {"statement": "I think that endangered species should be protected", "topic": "Environment"}

- {"statement": "The USA should increase funding to fight disease in developing nations", "topic": "Health"}

- {"statement": "We should punish parents who smoke in the presence of their children", "topic": "Health"}

- {"statement": "We should ban alcohol", "topic": "Health"}

- {"statement": "We should ban junk food from schools.", "topic": "Health"}

- {"statement": "This House Believes That Employees Should Be Compelled To Disclose Their HIV Status to Employers", "topic": "Health"}

- {"statement": "This House Believes that assisted suicide should be legalized", "topic": "Health"}

- {"statement": "We should use force to protect human rights abroad", "topic": "International"}

- {"statement": "We should expand NATO", "topic": "International"}

- {"statement": "I think democracy can be built as a result of interventions", "topic": "International"}

- {"statement": "I think sanctions should be used to promote democracy", "topic": "International"}

- {"statement": "I think parents should be able to choose the sex of their children", "topic": "Philosophy"}

- {"statement": "I think that the use of atomic bombs against Hiroshima and Nagasaki was justified", "topic": "Philosophy"}

- {"statement": "I think Sperm and egg donors should retain their anonymity", "topic": "Philosophy"}

- {"statement": "I think that Federal States are better than unitary nations", "topic": "Politics"}

- {"statement": "We should introduce positive discrimination to put more women in parliament", "topic": "Politics"}

- {"statement": "We should follow countries such as Senegal that have quotas for women in politics", "topic": "Politics"}

- {"statement": "I think all nations have a right to nuclear weapons", "topic": "Politics"}

- {"statement": "We should introduce recall elections.", "topic": "Politics"}

- {"statement": "We should negotiate with terrorists", "topic": "Politics"}

- {"statement": "We should lower the voting age to 16", "topic": "Politics"}

- {"statement": "We should legalize polygamy", "topic": "Religion"}

- {"statement": "We should allow gay couples to marry", "topic": "Religion"}

- {"statement": "We should support international adoption", "topic": "Society"}

- {"statement": "Governments should prioritise spending money on youth", "topic": "Society"}

- {"statement": "We should force the media to display, promote and report women\u2019s sport equally to men\u2019s sport", "topic": "Sport"}

- {"statement": "I think suicide should be a human right", "topic": "CMV"}

- {"statement": "The US should strictly enforce border security to prevent illegal entry", "topic": "CMV"}

- {"statement": "Drunk driving should not be a crime itself.", "topic": "CMV"}

- {"statement": "I don't think the duty of child raising should belong to the biological parents.", "topic": "CMV"}

- {"statement": "The fact that voting isn't mandatory is a good thing.", "topic": "CMV"}

- {"statement": "Gun - Control / Ban should not be implemented", "topic": "CMV"}

- {"statement": "No one over the age of 80 should be allowed to serve in goverment.", "topic": "CMV"}

- {"statement": "Hate Speech is Free Speech", "topic": "CMV"}

\\
    \bottomrule
	\end{tabular}
    \vspace{-2mm}
\caption{List of input propositions. 
} 
\label{fig:input_query}
\egroup
\vspace{-2mm}
\end{figure*}

\end{document}